# Analysis of Generalizability of Deep Neural Networks Based on the Complexity of Decision Boundary

Shuyue Guan
Department of Biomedical Engineering
The George Washington University
Washington DC, USA
frankshuyueguan@gwu.edu

Murray Loew
Department of Biomedical Engineering
The George Washington University
Washington DC, USA
loew@gwu.edu

*Abstract*— For supervised learning models, the analysis of generalization ability (generalizability) is vital because the generalizability expresses how well a model will perform on unseen data. Traditional generalization methods, such as the VC dimension, do not apply to deep neural network (DNN) models. Thus, new theories to explain the generalizability of DNNs are required. In this study, we hypothesize that the DNN with a simpler decision boundary has better generalizability by the law of parsimony (Occam's Razor). We create the decision boundary complexity (DBC) score to define and measure the complexity of decision boundary of DNNs. The idea of the DBC score is to generate data points (called adversarial examples) on or near the decision boundary. Our new approach then measures the complexity of the boundary using the entropy of eigenvalues of these data. The method works equally well for high-dimensional data. We use training data and the trained model to compute the DBC score. And, the ground truth for model's generalizability is its test accuracy. Experiments based on the DBC score have verified our hypothesis. The DBC is shown to provide an effective method to measure the complexity of a decision boundary and gives a quantitative measure of the generalizability of DNNs.

## I. INTRODUCTION

The generalization ability (generalizability) is an essential characteristic of classifiers in both machine learning and deep learning. A classifier with good generalizability performs well on unseen data. In most situations, a small portion of data taken from the training set as test/validation data is used to describe the generalizability. It would be valuable to analyze the generalizability of a classifier model directly, without test data, because it could help the model selection, and save time and data (data are precious in some cases) for training models.

A deep neural network (DNN) usually contains manymore parameters than training data. Based on traditional generalization analysis such as the VC dimension [1] or Rademacher complexity [2], DNNs tend to overfit the training data and demonstrate poor generalization. Much empirical evidence, however, has indicated that neural networks can exhibit a remarkable generalizability [3]. This fact requires new theories to explain the generalizability of neural networks. Two main approaches characterize studies of generalizability for deep learning [4]: a generalization bound on the test/validation error calculated from the training process [5], [6], and a complexity measure of models [7]–[9], motivated by the VC-dimension.

Classifiers overfitting the training data lead to poor generalizability. To limit the overfitting, several regularization techniques such as dropout and weight decay have been widely applied in training DNNs. As $\mathcal{L}_1$ and $\mathcal{L}_2$ regularization could generate sparsity for sparse coding [10], regularization techniques simplify the model's structure and then prevent the model from overfitting [11], [12]. This is because the simplified model cannot fit all training data precisely but must learn the approximate outline or distribution of the training data, which is the key information required to perform well (generalizability) on test data. On the other hand, the law of parsimony (Occam's Razor) [13] implies that any given simple model is *a priori* more probable than any given complex model [14]. Therefore, we hypothesize that, on a specific dataset, if two models have similar high training accuracy (close to 1), **the simpler model will have a higher test accuracy (better generalizability)**.

There are two ways to measure model complexity: 1) to examine trainable parameters and the structure of the model [8], [15]; 2) to evaluate the complexity of the decision boundary [16]–[18], which is the consequential representation of model complexity. Recently, an analysis of complexity of the decision boundary investigated adversarial examples that are near the decision boundary [19]–[21]. In this paper, for DNN models, we analyze generalizability based on the complexity of the decision boundary. Unlike other recent studies, we propose a novel method to characterize these adversarial examples to reveal the complexity of the decision boundary, and this method is applicable to datasets of any dimensionality.

## II. METHODS

### A. Adversarial examples

It is difficult to describe the decision boundary of a trained DNN model directly. Using adversarial examples is the key to this problem because they are near the boundary and could be considered as points sampled from the boundary. The boundary is described by these examples. Specifically, for a two-class classifier $f$, an adversarial example $x$ is one for which:

$$f(x) \approx 0.5$$

There are several approaches to generate the adversarial examples [19]–[21]; we apply a simple one [20] to linearly generate them. For example, for the two-class classifier $f$, as Fig. 1 shows, we select one training data point $a$ in Class 1 and another one $b$ in Class 2. The example $x$ on the line segment between $a$ and $b$ can be defined by:

$$x = \lambda a + (1 - \lambda)b, \ 0 \leq \lambda \leq 1$$

The line must cross the decision boundary because its two ends are in different classes. Hence, the adversarial example $c$ exists on the line.

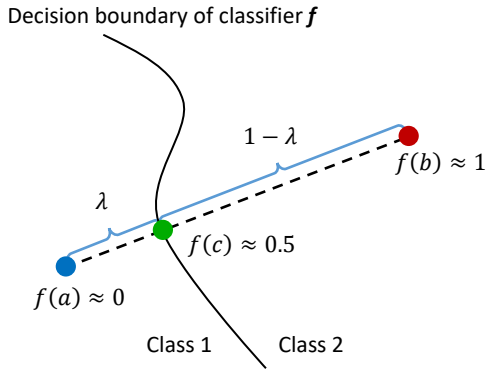

Fig. 1. To generate adversarial examples of classifier $f$

To find the adversarial example on such line segment, a simple method is to test different $\lambda$ from 0 to 1 by a small step. The pseudocode shows the algorithm of this process.

| **Algorithm** To find an adversarial example |
|---|
| 1:    $\forall \ a \in Class_1, \ b \in Class_2$    // $f(a) < 0.5; f(b) \geq 0.5$ |
| 2:    $c^* = a \ or \ b$    // initial the adversarial example |
| 3:    **for** $\lambda$ = 0 to 1 step $\varepsilon$ |
| 4:       $c = \lambda a + (1 - \lambda)b$ |
| 5:       // closer data point to decision boundary |
|        **if** $|f(c) - 0.5| < |f(c^*) - 0.5|$ |
| 6:          $c^* = c$ |
| 7:    **return** $c^*$ |

The precision of how the adversarial example closely locates to the boundary depends on the step ($\varepsilon$) value. And its time cost depends on the step ($\varepsilon$) too; it is about $\mathcal{O}(1/\varepsilon)$. This process can be speeded up to $\mathcal{O}(\log(1/\varepsilon))$ by the divide-and-conquer algorithm, which uses the **binary search**. In experiments, we set $\varepsilon = 1/256$.

*B. Boundary complexity measure*

For two-class datasets, one adversarial example is generated by a pair of data points from the two classes. Suppose Class 1 has $N$ data and Class 2 has $N$ data, randomly selected $N$ pairs of data can generate $N$ adversarial examples. As Fig. 2 shows, these adversarial examples (green points) are likely sampled from the decision boundary and could describe it. These generated adversarial examples form the **adversarial set**. We measure the complexity of the decision boundary by investigating the complexity of the adversarial set.

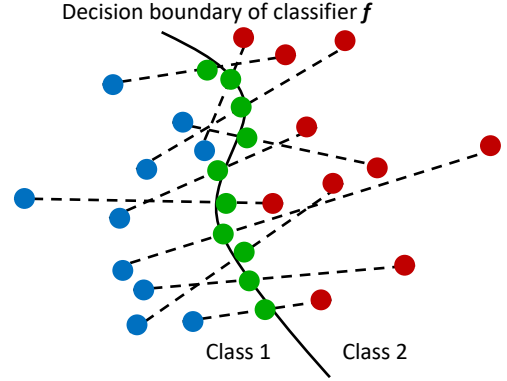

Fig. 2. Adversarial examples generated by pairs of data

We apply principal components analysis (PCA) to analyze the complexity of the adversarial set. In $n$-dimensions, an adversarial set with $m$ examples forms a $n \times m$ matrix $X$. Suppose $n < m$; by PCA, we have:

$$XX^TW = \lambda W$$

Where $W$ is eigenvector matrix: *s.t.* $W^TW = I$ and $\lambda$ contains the $n$ eigenvalues: $\{\lambda_1, \lambda_2, \cdots, \lambda_n\}$.

These **eigenvalues** could show the complexity of adversarial set. If $\frac{\lambda_i}{\sum \lambda_k} = 1$, it means all $m$ examples lie on the line of $i$-th eigenvector. It is the simplest condition for the adversarial set. If $\frac{\lambda_i + \lambda_j}{\sum \lambda_k} = 1$, it means all $m$ examples are on a plane; that indicates the decision boundary most likely is a plane. In general, we could measure the decision boundary complexity (DBC) of $f$ by computing the **Shannon entropy** of the eigenvalues:

$$\text{DBC}\{f\} = H\left\{\frac{\lambda_1}{\sum \lambda_i}, \frac{\lambda_2}{\sum \lambda_i}, \cdots, \frac{\lambda_n}{\sum \lambda_i}\right\}/\log n$$

Dividing by $\log n$ normalizes the DBC in range of [0, 1]. 0 is the simplest condition that the decision boundary is just a line.

A problem arises if we think about the most difficult condition of the boundary (DBC=1). For example, in 2-D, DBC=1 when the adversarial set forms a circle, but we cannot say the round boundary is the most complex one. For round-shape decision boundaries, some boundaries are smooth, and some may be lumpy. As Fig. 3 shows, the boundary (a) is more smooth (simpler) than (b). Through the hypothesis, we consider that the generalizability of model (a) is better than (b). However, DBC scores computed by adversarial sets of the two models will be similar (and close to 1).

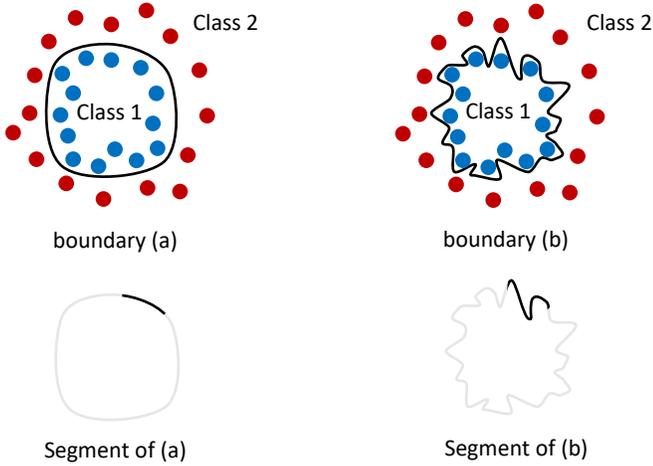

Fig. 3. Two kinds of round decision boundary

In Fig. 3, the boundary (a) is obviously simpler than boundary (b) because (b) has many zigzags in every segment. But if we compute the DBC score using the entire adversarial set, the effect (on eigenvalues' entropy) of zigzags is confused with the round-shape. Thus, it is not appropriate to use the entire adversarial set in such cases. If the adversarial set is generated by all data (in Fig. 2), we name it the **global adversarial set**. And the DBC score computed from it is called the **global DBC**. To solve the round-shape problem, we turn to consider adversarial examples on a section of the boundary, the segmental boundary. We define the adversarial data set formed by a segmental boundary as the **local adversarial set**.

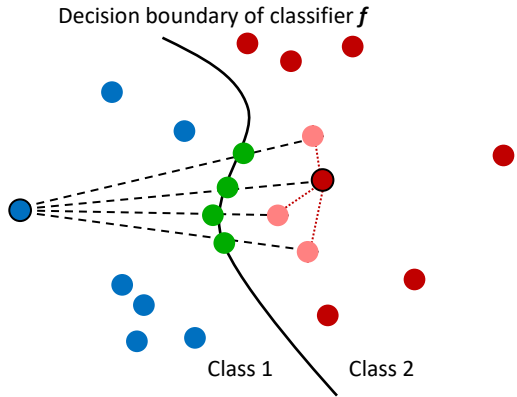

Fig. 4. Local adversarial set generated by 3-nearest neighbors of a pair

Adversarial examples in a local adversarial set should be close to each other to outline the shape of the segmental boundary. As Fig. 4 shows, a pair of data points from two classes is randomly selected, then to find $n$-nearest neighbors of one of those two data points. Finally, adversarial examples (green points) are generated by lines between these $n+1$ data points in one class to another data point in a different class. To decide the number of examples for one local adversarial set is an interesting question. It probably depends on the dimension and distances between example points. We will further discuss this question in Sec. III.B.

The computation process for complexity of a local adversarial set is the same as that for the global adversarial set. The steps show the process. The difference is that $N$ pairs of data generate one global adversarial set but $N$ local adversarial sets. Thus, one decision boundary has many **local DBC** scores.

| Steps | To compute one local DBC score |
|---|---|
| 1: | Take: $\forall\ a \in Class_1,\ b \in Class_2$ |
| 2: | Set of adversarial examples $\{c_1, c_2, \cdots, c_{k+1}\}$ generated by: $a$ to $k$-nearest neighbors of $b$ including $b$ |
| 3: | Compute the eigenvalues of the local adversarial set by PCA |
| 4: | Compute the normalized Shannon's entropy of eigenvalues |

## III. EXPERIMENTS AND RESULTS

We design three experiments to verify our boundary complexity measure. The dataset for the first experiment contains synthetic 2-D data with two classes. The second experiment uses the breast cancer Wisconsin dataset from *sklearn.datasets.load_breast_cancer*[1]. Its dimensionality is 30. The third experiment uses real images of cats and dogs downloaded from the GitHub[2]. The image size is 150x150x3; thus, these data are in very high dimension.

The key ideas of experiments are to train DNNs with different generalizabilities and compute DBC scores of these trained models. The ground truth for generalizability is the **test accuracy** because the final purpose of DNNs with high generalizability is to improve their performances on unseen data, which are represented by the test set. In the experiments the generalizability of a DNN is changed by intentional overfitting, such as by adding excessive trainable weights and removing regularization layers.

### A. Synthetic 2-D dataset

The dataset is generated by *sklearn.datasets.make_blobs*. The dataset has two well-separated clusters; each cluster has 200 data points and belongs to one class (see data points in Fig. 5), and thus this dataset is linearly separable.

Two fully-connected neural networks (FCNNs) have been trained to classify this dataset. There are no test data, and both training accuracies are 100%. Their real decision boundaries are shown in Fig. 5. Obviously, the decision boundary of model (a) is simpler than that of model (b) because to the linearly separable dataset, any non-linear boundary is superfluous.

For the two models (a) and (b), we generate adversarial examples (green points) by pairs of data from two classes to form the global adversarial sets, which clearly illustrate the boundary shape. And, the global DBC scores successfully show their different complexity situations (smaller DBC means simpler boundary). In this case, we could assert that the local DBC of model (a) must be smaller than the local DBC of model (b) without quantitative comparison because any segment of boundary (a) is not more complex than any segment of boundary

---

[1] https://scikit-learn.org/stable/modules/generated/sklearn.datasets.load_breast_cancer.html
[2] https://github.com/vyomshm/Cats-Dogs-with-keras

(b). Hence, the overall (average) local DBC of boundary (a) must be smaller than that of boundary (b). For convenience, in other experiments, we compute only the local DBC.

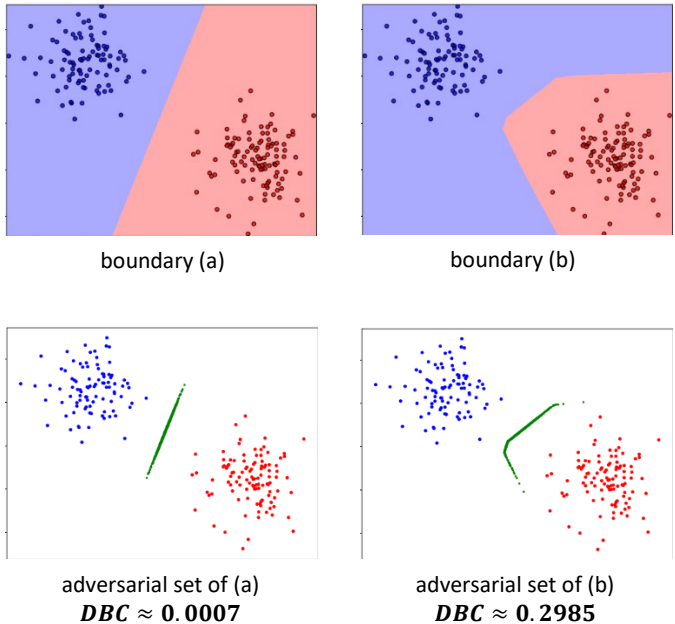

Fig. 5. Decision boundaries of two models trained by the synthetic 2-D dataset. The FCNN:(a) has only one hidden layer with one neuron; its number of parameters is 5 (including bias). The FCNN:(b) has three hidden layers with 10, 32 and 16 neurons; its number of parameters is 927 (including bias).

The DBC effectively detects the model with better generalizability (simpler decision boundary). It is not very impressive for the 2-D dataset because the boundary is visible. We could visually spot the simpler boundary case. But for a high-dimensional dataset, we must rely on the DBC score to describe the complexity of decision boundary.

### B. Breast cancer dataset

The dataset is imported from the breast cancer (Wisconsin) dataset and has two classes (212 Malignant and 375 Benign cases). Each case contains 30 numerical features.

Two FCNN models (bC1 and bC2) have been trained to classify this dataset. The training-test data ratio is 3:2 and both training accuracies are nearly 100% (>0.99) at the end. Then, we obtain models' test accuracies as the ground truth for model complexity. The greater test accuracy value means better generalizability (simpler decision boundary).

To compute the local DBC scores uses only the training data. We randomly select a pair of data points, of which one is a Malignant sample and another one is a Benign sample, to compute one local DBC score on trained models. This process is repeated 2,500 times (about 5 times of total number of data) to obtain 2,500 local DBC scores for each model. These local DBC scores are based on 30-nearest neighbors because the space dimension is 30. Thus, each local DBC score is computed by 31 adversarial examples. The reason is that, in 30-D, the simplest

---

[3] https://www.mathworks.com/help/stats/signrank.html

---

element (30-**simplex**) contains 31 vertices (*e.g.* as triangle in 2-D and tetrahedron in 3-D). We consider that $n$-nearest neighbors could best reflect the complexity of segmental boundary in $n$-D. The next experiment shows that the number of nearest neighbors could be much smaller than the dimensionality and not unique.

Fig. 6 and TABLE I clearly indicate that the model bC2 generally has bigger local DBC scores than bC1. The result means bC1 has better generalizability than bC2, which is verified by their test accuracies. We do not calculate the standard deviation of scores because their distributions are not Gaussian but more like the long-tailed distribution. Instead, we apply the two-sample rank test[3] to estimate whose scores are smaller.

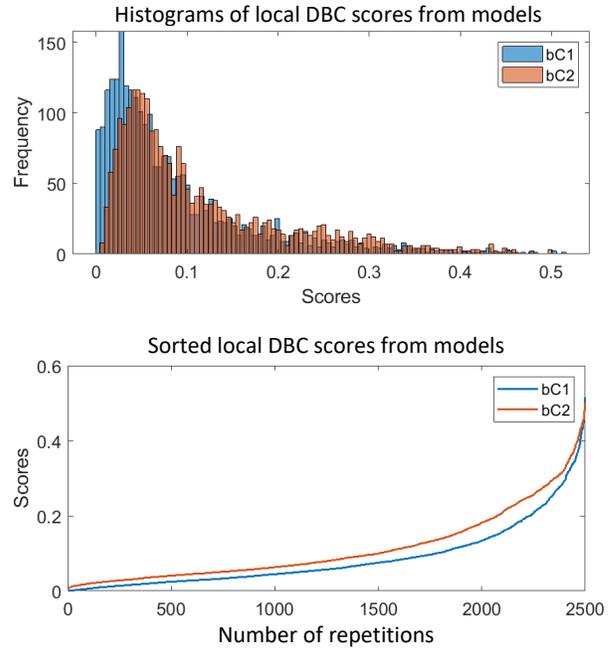

Fig. 6. Local DBC scores from two models trained by the breast cancer dataset. The FCNN bC1 has three hidden layers (20 neurons in each layer) and three Dropout layers; its number of parameters is 1,481 (including bias). The bC2 has one hidden layer with 1,000 neurons; its number of parameters is 32,001 (including bias).

TABLE I  STATISTICAL RESULTS OF LOCAL DBC SCORES ON bC1 AND bC2

| Model | Test Acc[a] | 2,500 local DBC scores | | |
|---|---|---|---|---|
| | | Mean | Median | h0 (bC1≥bC2)[b] |
| **bC1** | **0.970** | 0.087 | 0.057 | Rejected (p ≈ 0) |
| **bC2** | **0.921** | 0.113 | 0.079 | |

[a.] Test accuracy is the ground truth.
[b.] By Two-sample Wilcoxon signed rank test.

### C. Cat and dog dataset

This dataset contains 1,440 cat and 1,440 dog RGB photos. The image size is 150x150x3 (67,500 8-bit integers). Three convolutional neural network (CNN) models (cC1, cC2 and cC3) are trained to classify this dataset. The training-test ratio is 32:13 and both training accuracies are >0.95 at the end. Then, we obtain models' test accuracies as the ground truth for model complexity. Fig. 7 shows the training process.

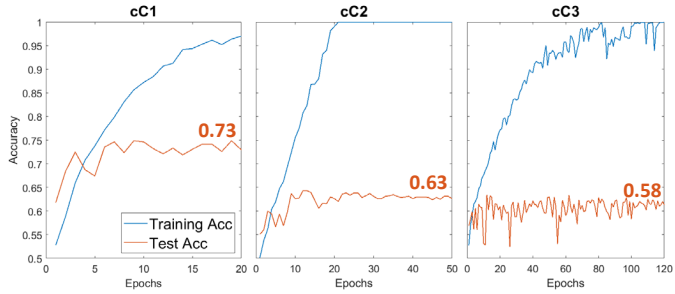

Fig. 7. Training and test accuracies in training process of three models. The CNN cC1 has three convolutional layers, three max-pooling layers, one dense layer (64 neurons) and one Dropout layer. The cC2 has one convolutional layer and three dense layers (256, 128, 64 neurons). The cC3 has only one dense layer (1024 neurons).

To compute the local DBC scores uses only the training set. We randomly select a cat and a dog image from the training set to compute local DBC scores on trained models. This process is repeated 6,000 times (about 5 times of the size of training set) to obtain 6,000 local DBC scores for each model.

Since the space dimension (67,500) is far beyond the size of dataset (2,880), we cannot choose based on the idea of a simplex and use 67,500-nearest neighbors to compute local DBC scores. Even if we have enough images to use, the number of nearest neighbors is too large to run the process. Hence, to find a properly small number, we test the 3, 5, 10, 15, 20, 30-nearest neighbors.

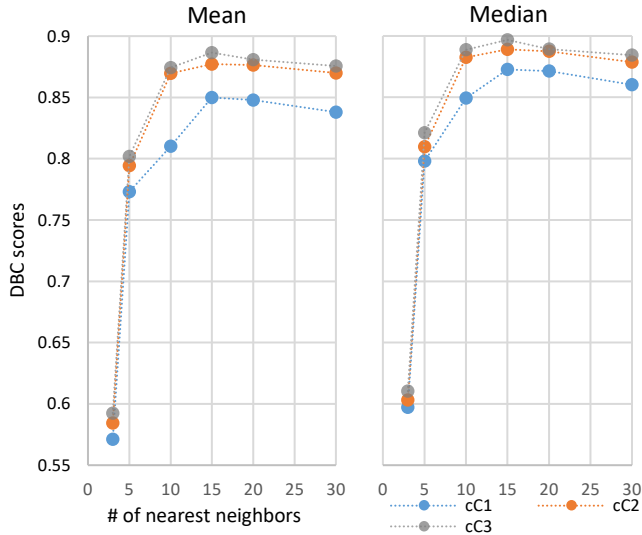

Fig. 8. Means and medians of local DBC scores on model cC1, cC2 and cC3 using different numbers of nearest neighbors.

Fig. 8 shows the means and medians of 6,000 local DBC scores based on various numbers of nearest neighbors. Since the distributions of these scores are not Gaussian but more like the long-tailed, we use their medians instead of the standard deviations. By comparing the means and medians for the three models, we find that, regardless of the number of nearest neighbors, for the DBC scores: cC1 < cC2 < cC3 always holds.

Such a conclusion is verified by their test accuracies (see test accuracy in TABLE II). The higher test accuracy suggests the model has better generalizability and should have a simpler decision boundary and smaller DBC scores. More strict estimates require the two-sample rank test. We provide an example of the 15-nearest neighbors case in TABLE II. We reject two null hypotheses: cC1≥cC2 and cC2 ≥ cC3 with p ≈ 0; it proves that cC1 < cC2 < cC3. Also, Fig. 9 indicates the same conclusion.

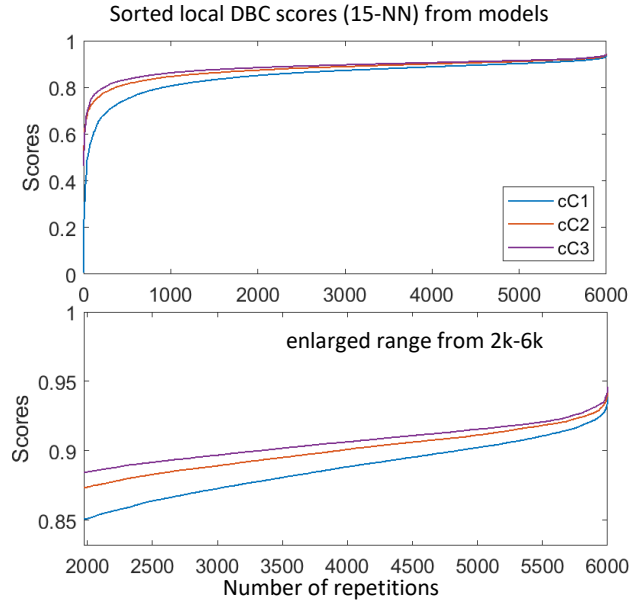

Fig. 9. Increasingly sorted local DBC scores from three models. The upper figure is the whole plot and the lower figure is zoomed the plot in range from 2k-6k to clearly see positions of three curves.

TABLE II  STATISTICAL RESULTS OF LOCAL DBC SCORES ON cC1, cC2 AND cC3

| Model | Test Acc[a] | 6,000 local DBC scores (15-NN[b]) | | | |
|---|---|---|---|---|---|
| | | Mean | Median | h0 (cC1≥cC2)[c] | h0 (cC2≥cC3) |
| cC1 | 0.730 | 0.850 | 0.873 | Rejected (p ≈ 0) | |
| cC2 | 0.626 | 0.877 | 0.889 | | Rejected (p ≈ 0) |
| cC3 | 0.583 | 0.887 | 0.897 | | |

[a.] Test accuracy is the ground truth.
[b.] Computation bases on 15-nearest neighbors.
[c.] By Two-sample Wilcoxon signed rank test.

## IV. DISCUSSION

The main idea of this study is simple and clear, that is, using the adversarial examples on or near the decision boundary to measure the complexity of the boundary. It is difficult to define and measure the complexity of a boundary surface in high dimensions, but easier to measure the complexity of adversarial example sets. We measure the complexity via the entropy of eigenvalues of adversarial sets. Other complexity measures for grouped data are also worth considering [22]. Fig. 10 shows several adversarial examples for the cC1 model generated by training images. They look like mixed cat and dog photos.

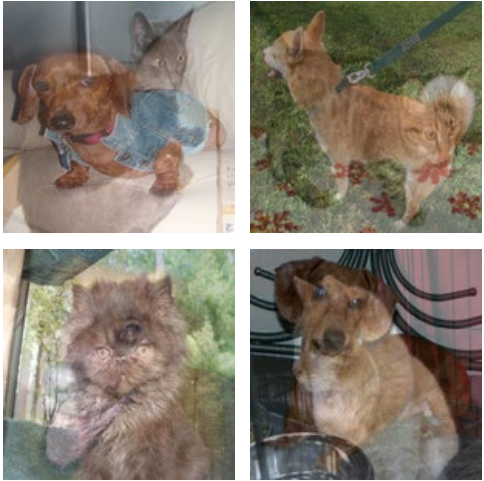

Fig. 10. Adversarial examples for the cC1 model

To generate the adversarial examples, as Fig. 1 shows, we use a pair of real data from different classes. At least one adversarial example is on the line segment between two data points because the line must cross the decision boundary at least once. If we use only real data from the training set, we could **evaluate a model's generalizability without using a test set**. That is an advantage when data are limited because we could have more data for training. However, the disadvantage of this method is the dependence on real data. The number of adversarial examples that could be generated depends on the size of the real dataset. Can we generate an adversarial example $x$ for classifier $f$ by randomly searching $f(x) \approx 0.5$? Maybe, but it is very difficult for the high-dimensional space. Even to find two data points $a, b$ whose $f(a) \approx 1, f(b) \approx 0$ is difficult because one of the areas (say $f(a) \approx 1$) would be very small and sparse in the space. Definitely, there are some other methods to generate adversarial examples, such as the DeepDIG [21] and applications of the generative adversarial network (GAN).

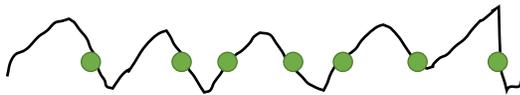

Fig. 11. Linear adversarial set on lumpy boundary

Smaller local DBC scores are **necessary but insufficient** conditions to simpler decision boundary because a lower complexity adversarial set may be generated from a higher complexity boundary (Fig. 11). Hence, the density of adversarial examples is important. Denser examples have higher probability to reflect the real condition of boundary. In practice, more adversarial examples are required to be on the effective segment of decision boundary, which is not the whole boundary but the part close to the data. From this aspect, to generate adversarial examples on the line segments between two data points is an appropriate way to create a dense adversarial set on the effective segment of decision boundary.

A smaller DBC score indicates that the model has a simpler decision boundary and better generalizability on a certain dataset. It is worth noting that the DBC score is meaningless for a single model and cannot be compared across different datasets. The gist of DBC score is used to compare various models **trained on the same dataset**. In this study, all three experiments use two-class datasets. In future work, we will use multiclass datasets. The multiclass problem could be treated as multiple two-class problems by one class vs. others.

## V. Conclusion

The decision boundary complexity of DNNs is defined and measured by DBC scores. This new DBC score is computed from the entropy of eigenvalues of adversarial examples, which are generated on or near the decision boundary, and in a feature space of any dimension.

Training data and the trained models are used to compute the DBC scores, and test data are used to obtain test accuracies as the ground truth for models' generalizability. Then, the DBC scores are used to detect the model with better generalizability trained on the same dataset. Results have verified our hypothesis that a DNN with a simpler decision boundary has better generalizability. In this paper, we provide an effective method to measure the complexity of decision boundaries and helps analyze the generalizability of DNNs.